\documentclass[letterpaper, 10 pt, conference]{ieeeconf}  

\IEEEoverridecommandlockouts                              

\usepackage{longtable}
\usepackage{graphicx}
\usepackage{amsmath,amssymb,amsfonts}
\usepackage[caption=false]{subfig}
\usepackage[ruled,linesnumbered, noend]{algorithm2e}
\usepackage{dirtytalk}
\usepackage{hyperref}
\usepackage{gensymb}
\usepackage{tabularx}
\usepackage{multirow}
\usepackage[short]{optidef}
\usepackage{booktabs}
\usepackage{siunitx}
\usepackage{multirow}
\usepackage{makecell}

\makeatletter
\newcommand{\removelatexerror}{\let\@latex@error\@gobble}
\newcommand\Tstrut{\rule{0pt}{2.0ex}}         
\makeatother
\SetKwComment{Comment}{$\triangleright$\ }{}

\title{\LARGE \bf
A real-time dynamic obstacle tracking and mapping system for UAV navigation and collision avoidance with an RGB-D camera}

\author{Zhefan Xu\footnotemark*, Xiaoyang Zhan\footnotemark*, Baihan Chen, Yumeng Xiu, Chenhao Yang, and Kenji Shimada 
\thanks{*The authors contributed equally.
    \newline{\indent Zhefan Xu, Xiaoyang Zhan, Baihan Chen, Yumeng Xiu, Chenhao Yang, and Kenji Shimada are with the Department of Mechanical Engineering, Carnegie Mellon University, 5000 Forbes Ave, Pittsburgh, PA, 15213, USA.}
        {\tt\small {zhefanx}@andrew.cmu.edu}}%
}

\begin{document}

\maketitle
\thispagestyle{empty}
\pagestyle{empty}

\begin{abstract}
The real-time dynamic environment perception has become vital for autonomous robots in crowded spaces. Although the popular voxel-based mapping methods can efficiently represent 3D obstacles with arbitrarily complex shapes, they can hardly distinguish between static and dynamic obstacles, leading to the limited performance of obstacle avoidance. While plenty of sophisticated learning-based dynamic obstacle detection algorithms exist in autonomous driving, the quadcopter's limited computation resources cannot achieve real-time performance using those approaches. To address these issues, we propose a real-time dynamic obstacle tracking and mapping system for quadcopter obstacle avoidance using an RGB-D camera. The proposed system first utilizes a depth image with an occupancy voxel map to generate potential dynamic obstacle regions as proposals. With the obstacle region proposals, the Kalman filter and our continuity filter are applied to track each dynamic obstacle. Finally, the environment-aware trajectory prediction method is proposed based on the Markov chain using the states of tracked dynamic obstacles. We implemented the proposed system with our custom quadcopter and navigation planner. The simulation and physical experiments show that our methods can successfully track and represent obstacles in dynamic environments in real-time and safely avoid obstacles. Our software is available on GitHub\footnote{\url{https://github.com/Zhefan-Xu/CERLAB-UAV-Autonomy}} as an open-source ROS package.

\end{abstract}

\section{Introduction}
Unmanned Aerial Vehicles (UAV) have been widely used in various fields \cite{construction}\cite{agriculture}\cite{zhefan_planner}\cite{rescue}, such as construction, agriculture, exploration, and rescue. The UAVs' working environments are usually challenging due to the complex static structures and unpredictable moving obstacles \cite{dpmpc}. For safe navigation in dynamic environments, an accurate and lightweight perception system is essential for tracking dynamic obstacles and mapping the static environments.  

There are mainly two categories of approaches for dynamic obstacle detection and tracking in UAV applications. The first category of  methods applies vision-based algorithms to detect obstacles \cite{Robust_avoidance_Umap}\cite{reactive_avoidance_Umap}\cite{ground_dynamic_avoidance}\cite{yolo_avoidance}. Some works \cite{Robust_avoidance_Umap}\cite{reactive_avoidance_Umap} use geometric information from a depth image to extract obstacles' bounding boxes and achieve real-time tracking performance. However, they cannot classify the static and dynamic obstacles. The others \cite{ground_dynamic_avoidance}\cite{yolo_avoidance} adopt the learning-based methods for obstacle avoidance with relatively higher computation resources. The second category of methods utilizes the pointcloud to build a voxel map and classify each voxel into static and dynamic. These methods can represent the dynamic environments efficiently, but their discrete voxel representation cannot accurately predict dynamic obstacles' future trajectories for collision avoidance.

\begin{figure}[t]
    \vspace{0.2cm}
    \centering
    \includegraphics[scale=0.345]{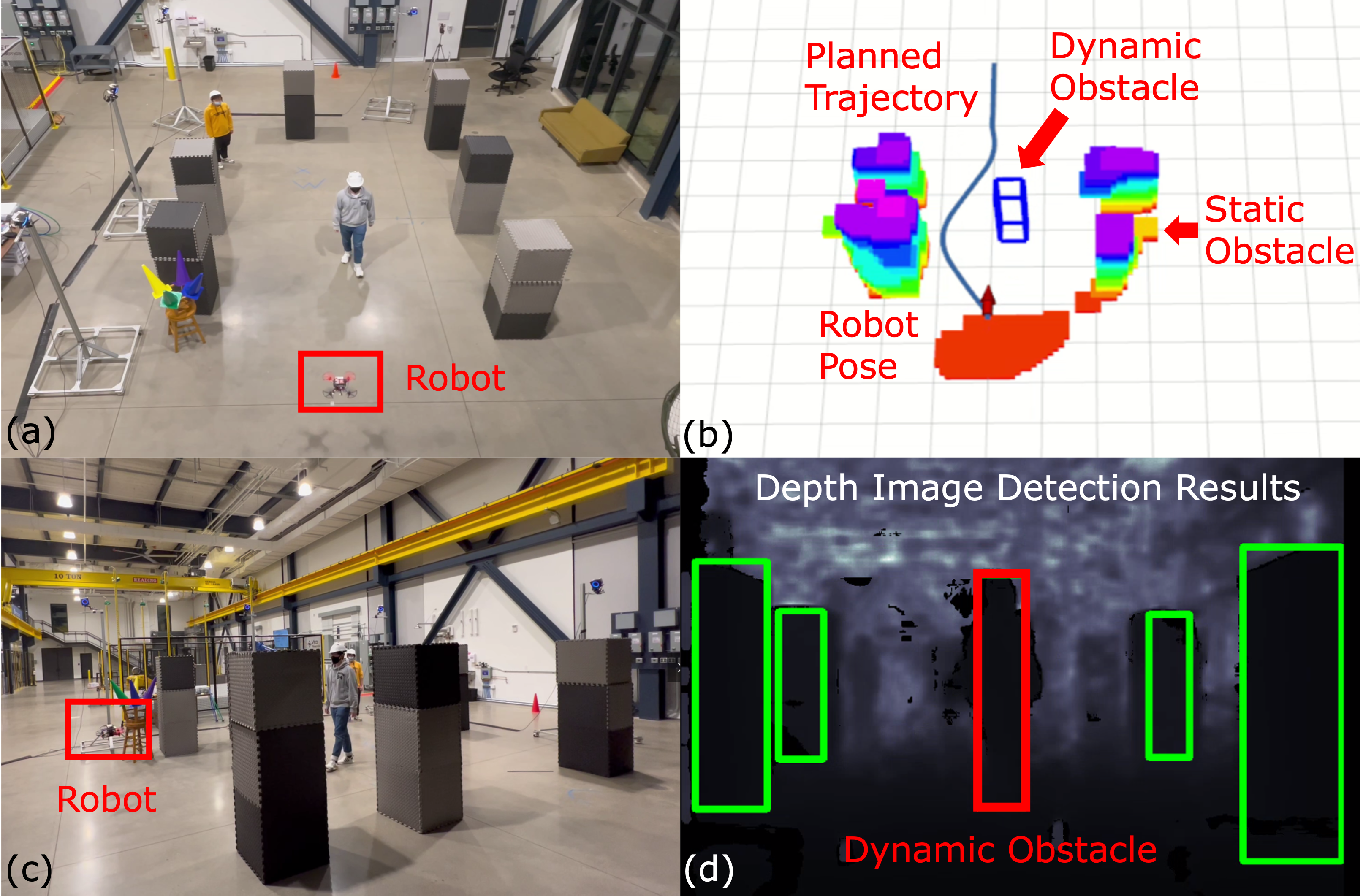}
     \caption{A physical flight experiment using the proposed mapping system. (a) A person walks toward the robot in the top view. (b) The robot generates the avoidance trajectory based on the detected dynamic obstacles and the voxel map. (c) The side view of the experiment. (d) The 2D bounding boxes of obstacles from the depth image during the flight.}
    \label{intro_figure}
\end{figure}

This paper introduces a novel real-time dynamic obstacle tracking and mapping system with the trajectory prediction module for dynamic environment navigation. The proposed system uses a 3D hybrid map that adopts an occupancy voxel map for static environment representation and tracks dynamic obstacles using their bounding boxes with velocity information. Our method applies the depth map-based detector to obtain 3D region proposals for dynamic obstacles and refine the bounding boxes' sizes by our occupancy map refinement approach. Then, we apply the Kalman filter and our continuity filter to identify and track dynamic obstacles. Next, the proposed dynamic-region cleaning method is applied to free the moving obstacle areas in the static map. Besides, the Markov chain-based trajectory prediction method is developed for dynamic obstacles, considering the interaction between dynamic obstacles and static environments. The contributions of this paper are as follows: 
\begin{itemize}
    \item \textbf{Region proposal detector with map refinement:} The proposed method applies a lightweight depth image detector to obtain obstacle region proposals and uses the static map to refine the obstacles' bounding boxes.
    \item \textbf{Dynamic obstacle identification and tracking:} We apply the Kalman filter and our continuity filter to track and identify dynamic obstacles. The dynamic-region cleaning approach is then applied to clean the remained trails of dynamic obstacles in the static map.
    \item \textbf{Environment-aware trajectory prediction:} The proposed Markov chain-based dynamic obstacle trajectory prediction method considers the interaction between dynamic obstacles and the static environment based on the trajectory probability distribution.
\end{itemize}

\section{Related Work}
Recent years have seen plenty of obstacle detection and tracking algorithms based on different sensors, such as LiDAR \cite{LIDAR}\cite{Falcon}, monocular camera \cite{monocular1}\cite{monocular2}, event camera \cite{event_camera}\cite{eventcamera_multivehicle}, and depth camera \cite{Robust_avoidance_Umap}\cite{reactive_avoidance_Umap}. The depth camera, one of the most popular sensors for small-size UAV navigation, can provide the robot with images and pointcloud data. Based on the input data representations, there are two categories of obstacle detection and tracking methods:

\textbf{Vision-based method:} In \cite{Robust_avoidance_Umap}\cite{reactive_avoidance_Umap}, the u-depth maps are extracted from the depth images to detect obstacles in the environments. The results from \cite{Robust_avoidance_Umap} show successful obstacle avoidance in dynamic environments. However, static obstacles are represented using the ellipsoid, which can be overly-conservative when the shape of a static obstacle becomes complex. \cite{feature_avoidance} applies a 2D feature-based perception method to achieve obstacle avoidance. \cite{ground_dynamic_avoidance} adds a learning-based detector to improve the detection accuracy, and \cite{yolo_avoidance} applies the neural network to extract features, helping both perception and decision making. However, introducing learning-based methods makes the whole process slower and more computationally demanding \cite{YOLO_Vehicle}\cite{UAV_small_object}. 

\textbf{Pointcloud-based method:} This category utilizes the pointcloud data to build a dynamic map for both static and dynamic obstacles \cite{ground_dynamic_avoidance}\cite{ZJU_dynamic_avoidance}\cite{rebuttal_SJU}\cite{kernal_dynamic_map}\cite{sju_map_and_tracking}.  \cite{ground_dynamic_avoidance} applies the pointcloud clustering, using a ``per-point velocity'' voting strategy to identify dynamic obstacles. Inspired by \cite{ground_dynamic_avoidance}, \cite{ZJU_dynamic_avoidance} further fuses the dynamic obstacle detection results into a static map generated from the depth pointcloud. \cite{rebuttal_SJU} represents dynamic and static obstacles in a particle map, capturing the complicated shapes of dynamic obstacles. \cite{kernal_dynamic_map} uses kernel inference in the mapping process to reduce the computation. \cite{sju_map_and_tracking} leverages the voxel map to identify each voxel into static and dynamic and estimate the velocities of dynamic voxels.

Most of the works mentioned earlier are either computationally demanding or unable to represent complicated static structures with dynamic obstacles simultaneously. Besides, some methods only apply linear propagation for obstacle trajectory prediction, resulting in the suboptimal performance of obstacle avoidance. To address these issues, we propose a real-time dynamic obstacle tracking and mapping system with the trajectory prediction module for dynamic environment navigation. Similar to \cite{Markov2}\cite{Markov1}\cite{Markov3}, we adopt the Markov chain and introduce environment impact to trajectory prediction, inspired by the idea of social force \cite{socialforce1}\cite{socialforce2}.

\begin{figure}[t]
    \vspace{0.2cm}
    \centering
    \includegraphics[scale=0.175]{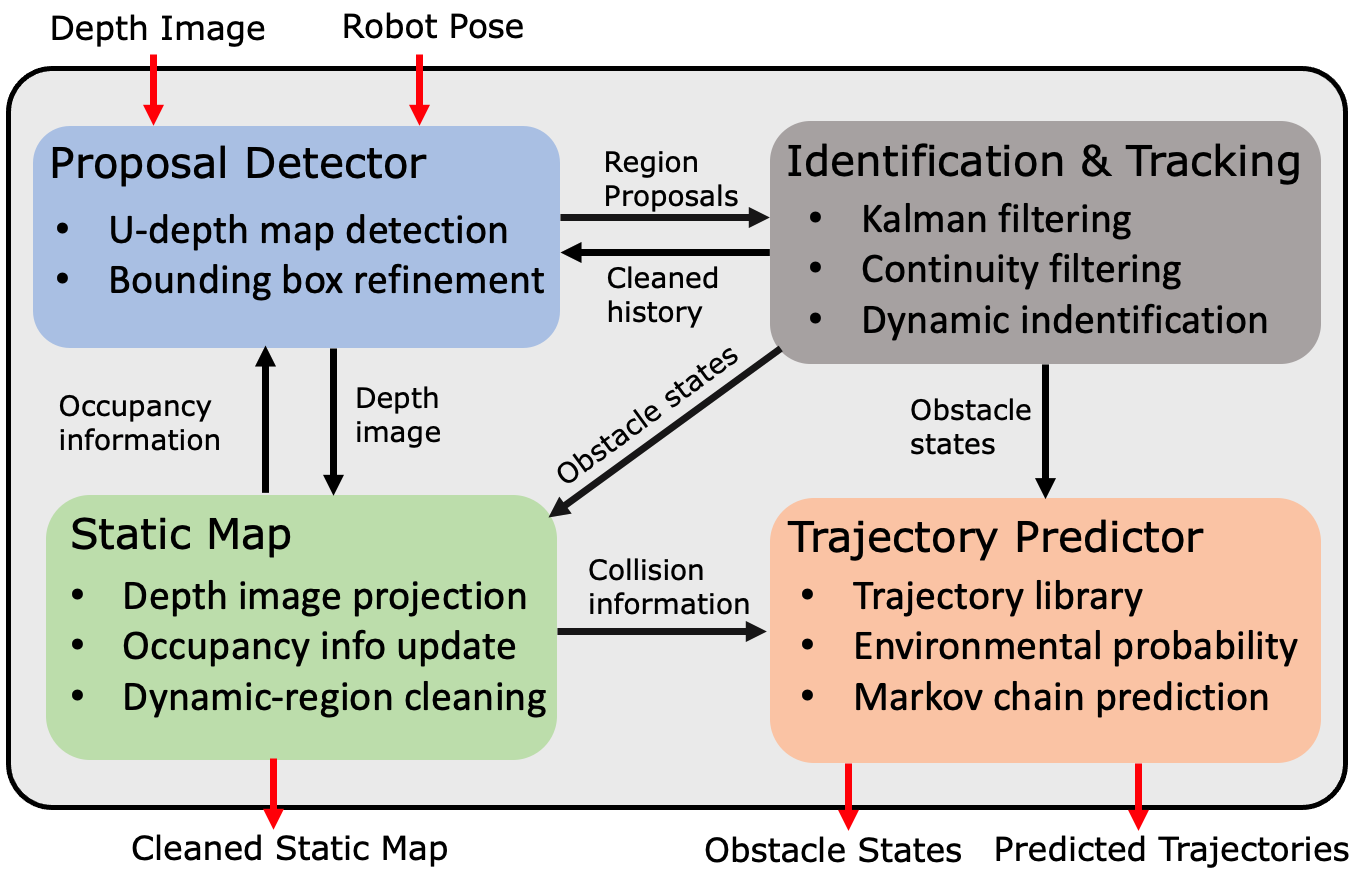}
    \caption{System framework. The system contains three dynamic modules with one static occupancy map module. It takes the depth images with the robot poses as the inputs and outputs cleaned static maps and dynamic obstacle states with their predicted trajectories.}
    \label{fig:pipeline}
\end{figure}

\section{Methodology}

\subsection{System Framework} 
The whole system can be split into three dynamic modules and one static map module, shown in Fig. \ref{fig:pipeline}. First, the \textbf{region proposal detector} detects static and dynamic obstacles and fuses them with the static map to refine the obstacles' region proposals. Then, the \textbf{dynamic obstacle identification and tracking} module tracks the history of obstacles' states and applies the Kalman filter and our continuity filter to identify the dynamic obstacles. Next, the states of dynamic obstacles are fed into the \textbf{static map} to clean the voxel residuals caused by the dynamic obstacles' motions. In the last module, \textbf{trajectory predictor}, the environment-obstacle interaction is introduced into calculating the environment probability for the Markov chain-based trajectory prediction. 

\subsection{Obstacle Region Proposal Detection}
This section describes depth image-based obstacle detection and obstacle region proposal generation with our map refinement method.

\begin{figure}[t]
    \vspace{0.2cm}
    \centering
    \includegraphics[scale=0.35]{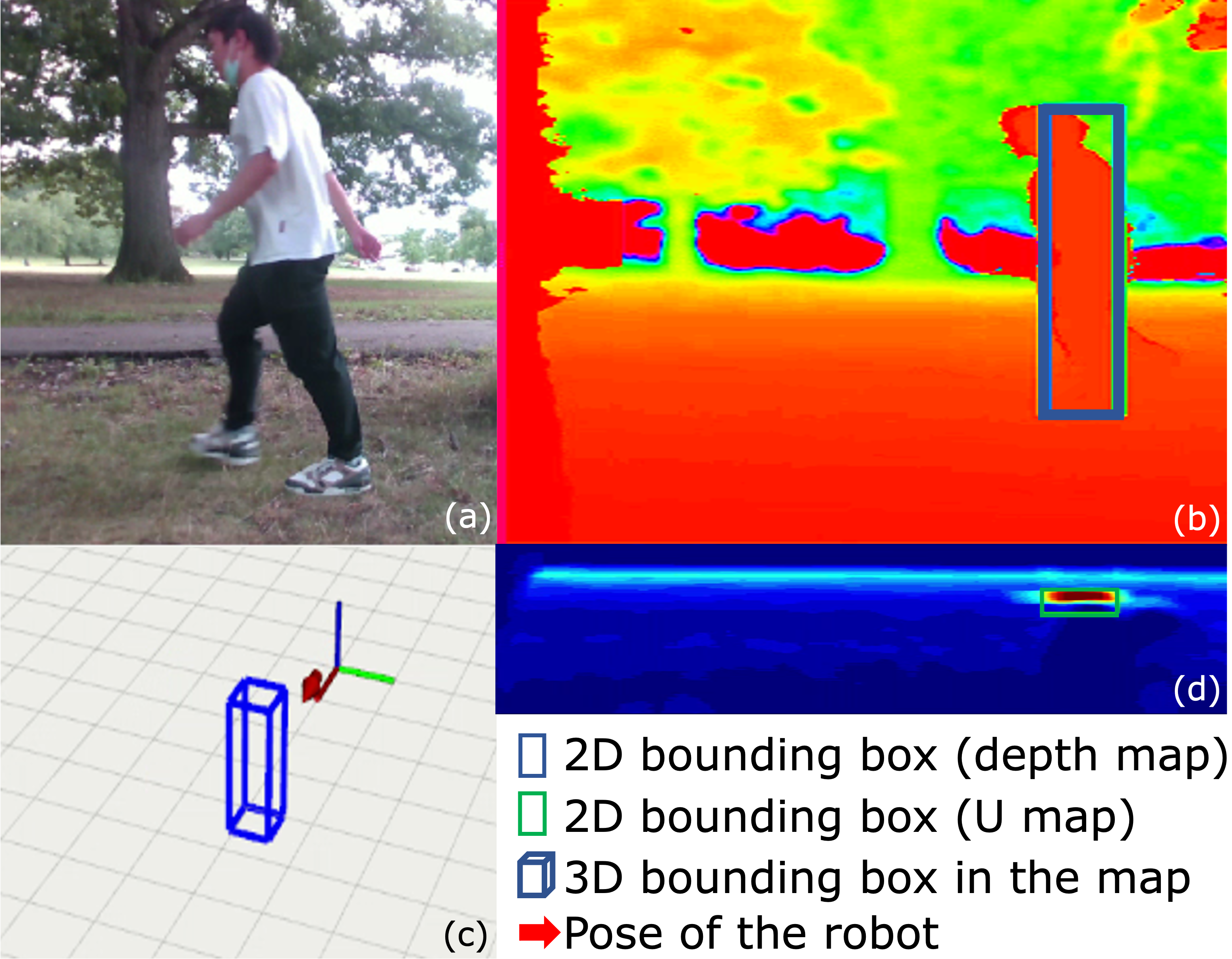}
    \caption{Illustration of detecting the raw bounding box of obstacles. (a) The RGB image. (b) The depth map with a 2D detected obstacle. (c) The 3D bounding box in the map frame. (d) The U-map with the 2D bounding box.}
    \label{fig:woods}
\end{figure}

\textit{1) Obstacles detection:} This part generates the ``raw'' 3D bounding box results of obstacles, which contain the rough positions and sizes of static and dynamic obstacles. The raw detection results are generated from the U map inspired by \cite{Robust_avoidance_Umap}\cite{reactive_avoidance_Umap}. A U map is computed with the column depth value histograms of the original depth image \cite{Robust_avoidance_Umap}. Assuming obstacles have continuously changing depth values in the depth images, which differ dramatically from the background, we can get the 2D boxes of the obstacles in the U map, as shown in Fig. \ref{fig:woods}d, by grouping regions of pixels with values higher than a threshold. This process can provide us with the rough obstacles' lengths and widths. Similarly, the heights on the depth image could be obtained by searching for pixels with depth values close to the corresponding U map 2D box on each column shown in Fig. \ref{fig:woods}b. So, a box on the image frame can be represented by its center ${P}^{I}_{o} = [{x}^{I}_{o},{y}^{I}_{o}]^{T}$ and the size $S^{I}_{o}=[{w}^{I}_{o},{h}^{I}_{o}]^{T}$ with the depth $d$, where $I$ denotes the image frame and $o$ denotes the obstacle. The bounding box could then be projected on the camera frame, and further transformed to the map frame with the center of ${P}^{M}_{o} = [{x}^{M}_{o},{y}^{M}_{o}, d_{o}]^{T}$ and size $S^{M}_{o}=[{w}^{M}_{o},{h}^{M}_{o},l^{M}_{o}]$, where $M$ represents the map frame. Fig. \ref{fig:woods}c gives a visualization.

\begin{figure}[t]
    \vspace{0.2cm}
    \centering
    \includegraphics[scale=0.155]{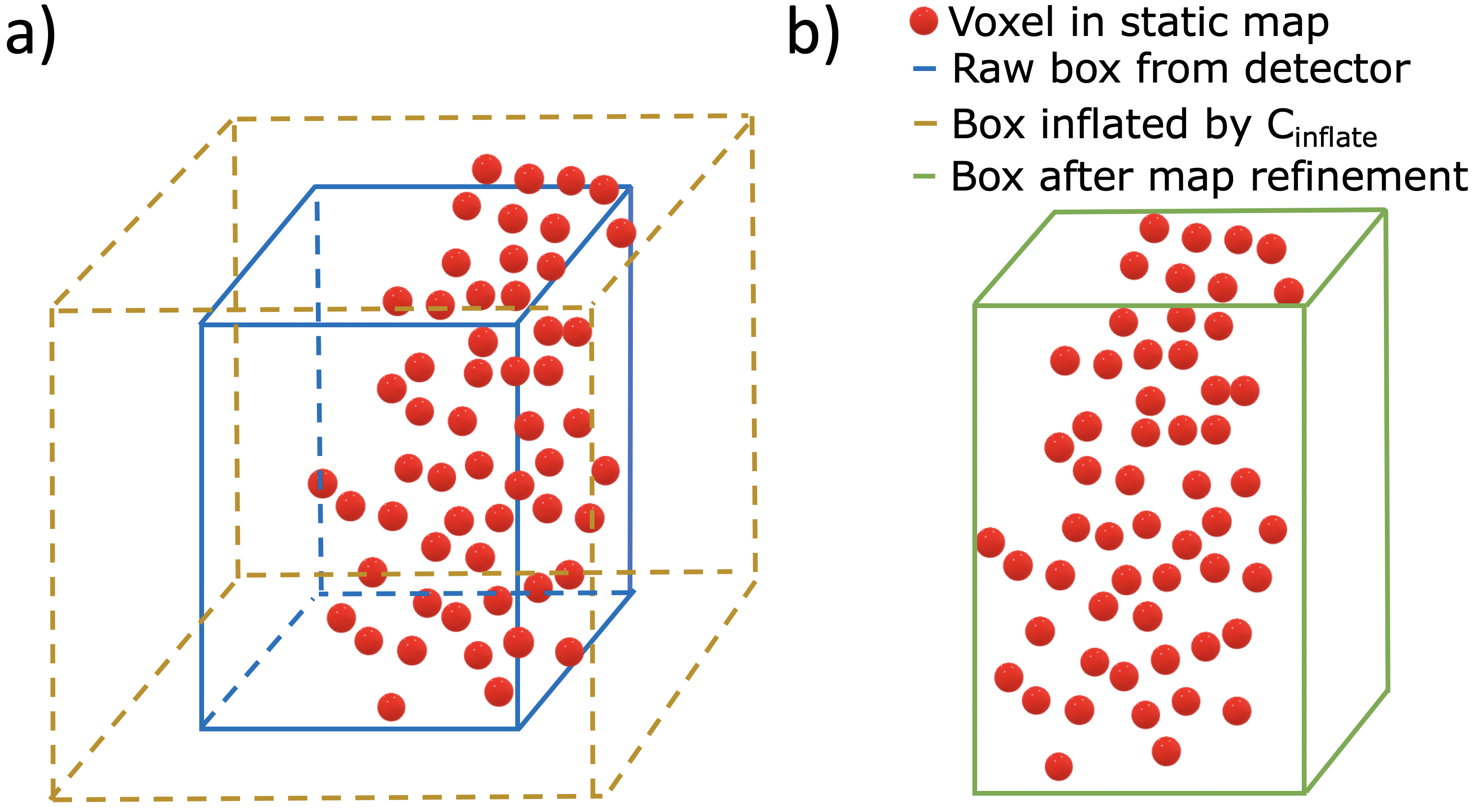}
    \caption{Illustration of the map refinement. a) The red points represent the voxel in the static map, and the blue box is the raw box generated from the detector, which might have a misestimated shape and size. The brown box is inflated by $C_{inflate}$ b) After searching for occupied voxels, the green box, which is a refined box, gives an improved result.}
    \label{fig:depth_cloud}
\end{figure}

\textit{2) Map refinement:}
Since the raw detection results from the previous step are not accurate enough due to the noises from the depth images, we refine the detected raw bounding boxes with the static occupancy map to improve the robustness and accuracy. This strategy is named as \textbf{map refinement}. As shown in Fig. \ref{fig:depth_cloud}a, we inflate the above-derived raw bounding boxes in the x, y, and z directions by multiplying a user-defined coefficient factor, $C_{inflate}$. Then, we search for the occupied voxels in the static occupancy map inside the proposal regions to get the minimal boxes that contain all occupied voxels as the final dynamic obstacle region proposals shown in Fig. \ref{fig:depth_cloud}b. 

\subsection{Dynamic Obstacle Identification and Tracking}
This module tracks the obstacle states with the histories, identifies dynamic obstacles, and cleans the occupied voxels on the trails of dynamic obstacles in the static map. The obstacle states contain the position, velocity and size data.

\textit{1) Obstacle tracking:}
For obstacle tracking, all obstacles must be associated with their previous positions in previous time frames. For the detected obstacles set $^{t}O^{C} = \left\{^{t}o^{C}_{0},^{t}o^{C}_{1}  ,\cdots,^{t}o^{C}_{n}\right\}$ in the camera frame at time $t$, we associate each bounding box $^{t}o^{C}_{i}$ with its best match $^{t-1}o^{C}_{j}$ at the previous time $t-1$. The match criteria is based on the closest bounding box center distance with the requirement of a high bounding box overlap ratio $r_{i,j}$:
\begin{equation}
    r_{i,j} = \frac{A_{i,j}}{A_{i}},
\end{equation}
where $A_{i,j}$ is the top-view overlap area between $^{t}o^{C}_{i}$ and $^{t-1}o^{C}_{j}$, and $A_{i}$ is the area of $^{t}o^{C}_{i}$.
By associating the previous $k+1$ time step's states, we obtain a tracking history for each obstacle $^{t}H_{i,k}^{C} = \left\{^{t}o_{i}^{C},^{t-1}o_{i}^{C},\cdots,^{t-k}o_{i}^{C}\right\}$. To reduce the effects of partially observed bounding boxes, we fix the bounding box sizes once the entire obstacles show up in the camera's field of view (FOV) for $k^{'}$ frames. This strategy is found to be useful for small FOV cameras.

\begin{figure}[t]
    \centering
    \includegraphics[scale=0.22]{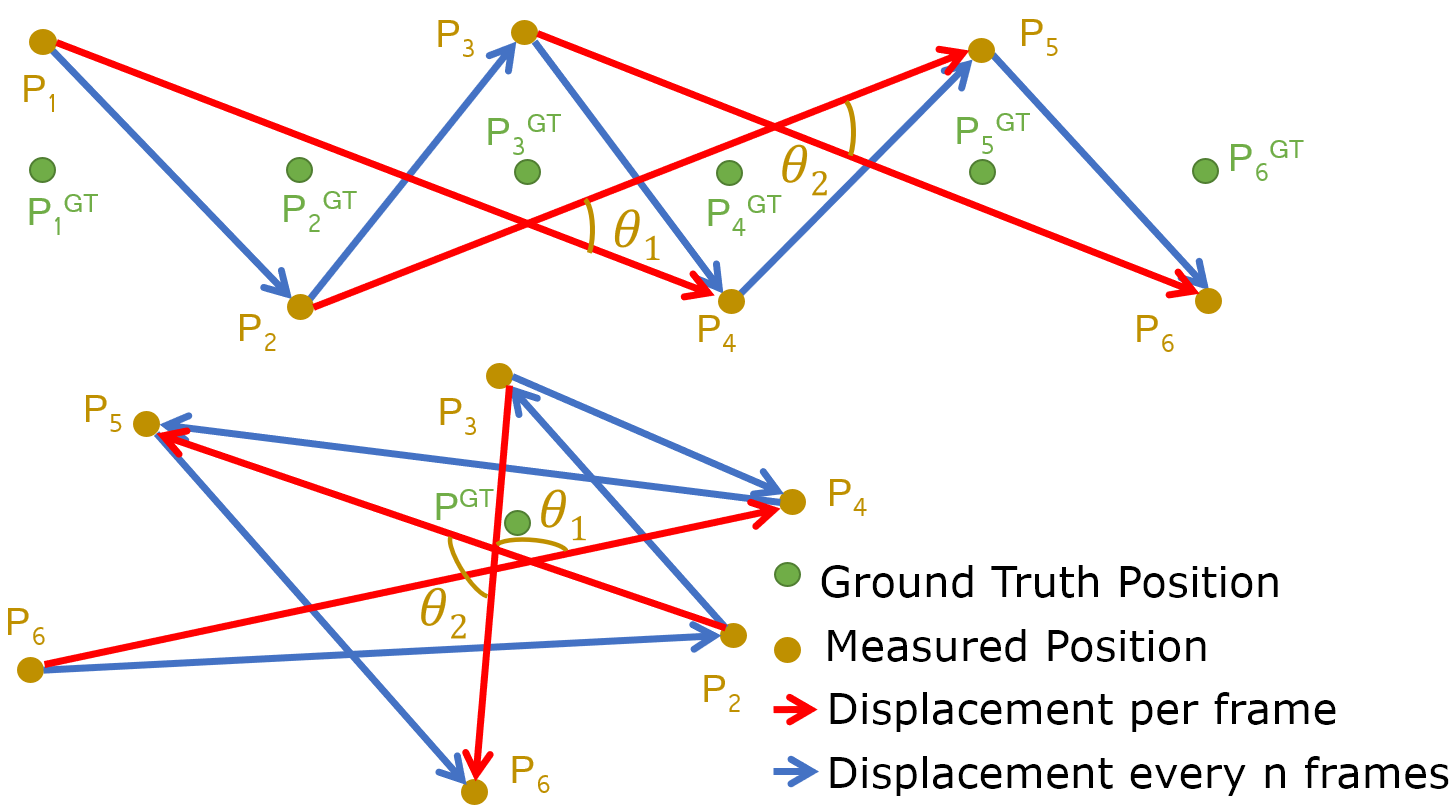}
    \caption{Illustration of the continuity filter. The k=6 frames' history of positions of a dynamic obstacle and a static obstacle are shown in the top and bottom, respectively. When the obstacle is dynamic, the angles $\theta_{1}$ and $\theta_{2}$ are smaller.}
    \label{fig:point_position}
\end{figure}

\textit{2) Identify dynamic and static obstacles:}
We apply the Kalman filter in the map frame to estimate the obstacles' velocities. For simplicity, we assume all variables are in the map frame. We denote the obstacle state vector as $X = [x, y, \Dot{x} , \Dot{y}]^{T}$ and the measurement vector as $Z = [o_{x}, o_{y}, v_{x} , v_{y}]^{T}$, where $v_{x}$, $v_{y}$ are velocities in the x and y directions calculated by each obstacle's latest two states. The system's  model and measurements are described by:
\begin{equation}
    X_{t|t-1} = AX_{t-1} +Q
    ,
\end{equation}
\begin{equation}
    Z_{t} = HX_{t} + R 
    ,
\end{equation}
where $A$ is the state transition model, $Q$ is the covariance of the model noise, $H$ is the measurement model, and $R$ is the covariance of measurement noise.
With the estimated velocity from the Kalman filter, we can identify a detected obstacle as dynamic if its velocity exceeds the user-defined threshold. However, some static obstacles might be incorrectly identified as dynamic due to the sensor measurement noises. To solve this issue, we propose the \textbf{continuity filter} to filter out incorrectly identified static obstacles. In Fig. \ref{fig:point_position} top, 6 frames of dynamic obstacle tracking histories are recorded. The brown points and blue arrows are the measured positions and trajectories of a dynamic obstacle. The green points are the ground truth positions. The filter first calculates the displacement vectors (red arrows) using the obstacle pairs from the tracking history defined as:
\begin{equation}
    Pa=\left\{(o_{i}^{t-n},o_{i}^{t})_{0},\cdots, (o_{i}^{t-k},o_{i}^{t-k+n})_{k-n}\right\}.
\end{equation}
Then, the displacement vectors between each point in the pairs are defined as:
\begin{equation}
D=\left\{d_{0}^{t-n,t},d_{1}^{t-n-1,t-1},\cdots,d_{k-n}^{t-n+k,t-k}\right\},
\label{eqn:displacement}
\end{equation}
where $d_{0}^{t-n,t}$ means the distance between point $t-n$ and point $t$.
After that, the cosine values of the angles between each pair of vectors with consecutive indices are obtained:
\begin{equation}
    \Theta = \left\{cos(\theta_{0,1}), cos(\theta_{1,2}),\cdots,cos(\theta_{k-n-1,k-n})\right\}.
\end{equation}
Assuming moving obstacles move with a constant velocity for a short period, the angles between displacement vectors should be close to 0, as the top figure in Fig. \ref{fig:point_position} shows. Conversely, the bottom figure in Fig. \ref{fig:point_position} shows that a static obstacles' measurement generates large angles between displacement vectors. Therefore, we define the \textbf{continuity coefficient} as:
\begin{equation}
    C_{con} = \frac{\Sigma\Theta}{k-n+1}.
\end{equation}
The detected obstacles with $C_{con}$ less than a threshold, $T_{con}$, will be marked as dynamic temporarily. Then, we keep track of $c$ frames of voting histories, which record the temporary identification result of each obstacle over the past $c$ frame. All the previous result votes for the classification of the current frame, and only those candidates with more than $T_{c}$ vote as dynamic will be finally determined as dynamic.

\textit{3) Dynamic-region cleaning:} Since the dynamic obstacles should not be contained in the static map, it is necessary to clean their motion histories in the static map, and this process is named as \textbf{dynamic-region cleaning}. Simply removing the occupancy voxels on the current dynamic obstacles' position is not enough because it may cause later bounding boxes in the next iteration to fail to search for occupied voxels during the map refinement. Therefore, we first clean the occupancy voxels based on the positions of dynamic obstacles in the past $f$ frames. Then, we record the removed voxels into a hash table known as the clean history in each iteration, and the voxels in clean history would be recognized as occupied during map refinement in the next iteration. 

\begin{figure}[t]
    \centering
    \includegraphics[scale=0.23]{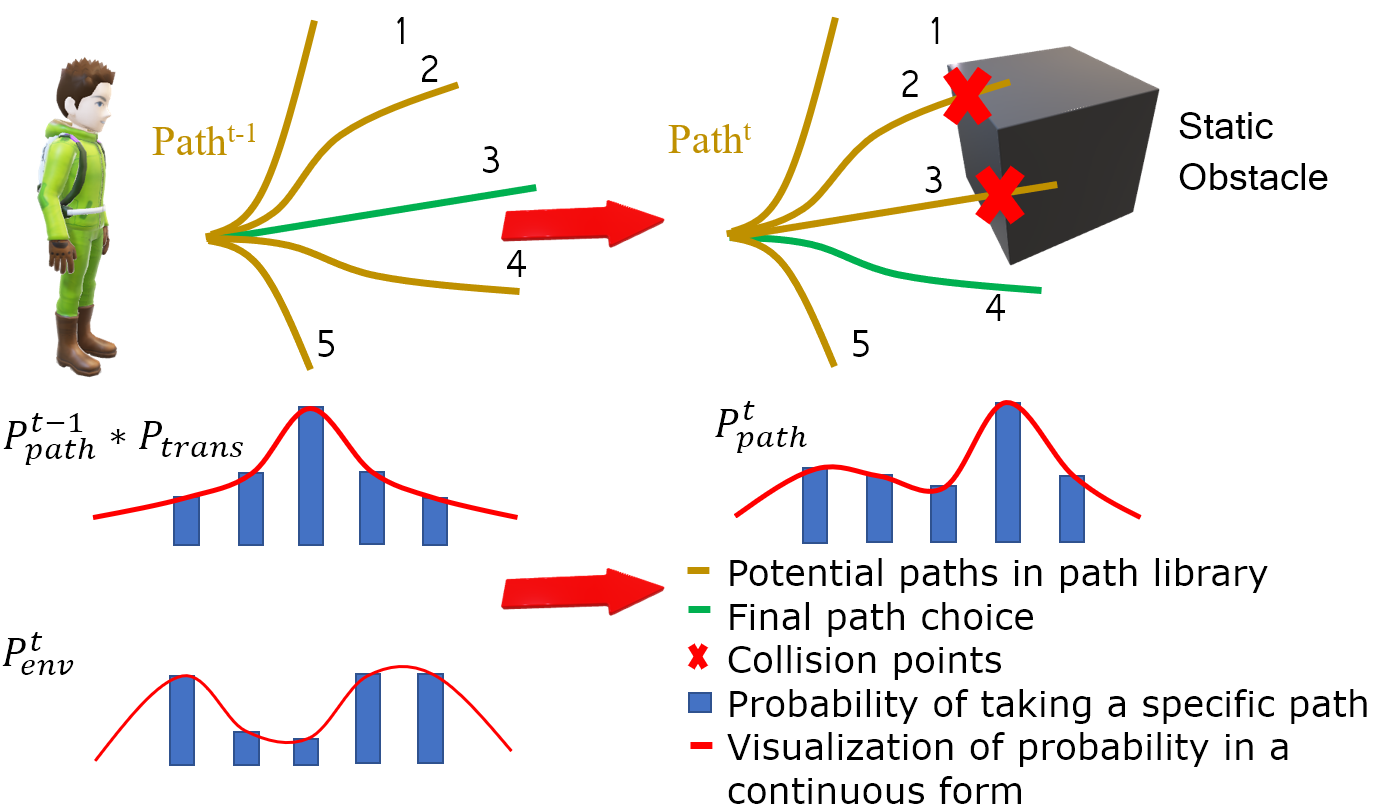}
    \caption{Illustration of the trajectory prediction. At the time $t-1$, all five paths in the path library are safe, so path 3 has the highest probability. However, at time $t$, the previously safe paths 2,3 in the path library are not safe anymore. So, the final prediction is path 4.}
    \label{fig:trajectory_prediction}
\end{figure}

\subsection{Trajectory Prediction}
This module applies an environment-aware Markov chain \cite{Markov2}\cite{Markov1}\cite{Markov3} to predict obstacle trajectories. The whole process is illustrated in Fig. \ref{fig:trajectory_prediction}. First, we generate a library of $l$ possible paths by collecting human walking data in experiments and fitting them with polynomials. The path library vector starts with the left-turn paths and ends with the right-turn paths with the straight path at the center. Then, the most likely path is chosen by calculating the probability distribution over $l$ paths, $P_{path}$. To calculate the probability distribution over paths, the initial state in the Markov chain are defined as the probability of each path:

\begin{equation}
    P_{init} =
\begin{bmatrix}
 p_{init}^{0} & p_{init}^{1} & \cdots & p_{init}^{l-1}
\end{bmatrix},
\end{equation}
where all values obey a discrete Gaussian distribution, and the probability values are obtained from Gaussian kernels with $p_{init}^{\frac{l}{2}}$ as the mean because $p_{init}^{\frac{l}{2}}$ is the path of going straight, and we assume a person always tends to choose paths close to a straight line if possible. In addition, we define the transition matrix in the Markov chain as:
\begin{equation}
    P_{trans} = 
\begin{bmatrix}
 p_{trans}^{0,0} & p_{trans}^{0,1} & \cdots &p_{trans}^{0,l-1}\\
p_{trans}^{1,0} & p_{trans}^{1,1} & \cdots &p_{trans}^{1,l-1}\\
\vdots & \vdots & \ddots & \vdots \\
p_{trans}^{l-1,0} & p_{trans}^{l-1,1} & \cdots & p_{trans}^{l-1,l-1}
\end{bmatrix},
\end{equation}
where each row is a discrete Gaussian distribution from Gaussian kernels with $p_{trans}^{i,i}$ as the mean since people tend to keep their moving tendencies whenever possible. To consider the environment-obstacle interactions, for each path in the library, we calculate the distance from its start to the collision point with the static map as $Dist_{i}$. Then, we feed the distances into a softmax function to calculate the \textbf{environment probability}, which describes the probability of choosing a specific path when considering the environment-obstacle interactions:

\begin{equation}
    P_{env} = \textbf{\emph{Softmax}}(
    Dist_{0}, Dist_{1}, ..., Dist_{l-1}),
\end{equation}
Finally, the state is predicted by :
\begin{equation}
    P_{path}^{t+1} = P_{path}^{t}P_{trans}*P_{env}^{t+1},
\end{equation}
where $*$ represents the elementwise multiplication.

\section{Result and Discussion}
\subsection{Implementation details}
To evaluate the performance of the proposed method, we conduct experiments in simulation and physical environments. The simulation experiments are implemented in C++ with ROS/Gazebo running on AMD Ryzen 7 5800@3.4GHz. For physical experiments, the system runs on our customized quadcopter with an Intel Realsense D435i depth camera, which provides 640 $\times$ 480 pixels depth images with 87\textdegree {} by 58\textdegree {} field of view. The Nvidia Xavier NX is used for onboard computation, and the PX4-based flight controller controls the robot in flight tests. We apply the visual-inertial odometry (VIO) \cite{qin2018vins} to estimate the states of the robot.

\subsection{Simulation and Physical Experiments}
\begin{figure}[t]
    \centering
    \includegraphics[scale=0.9]{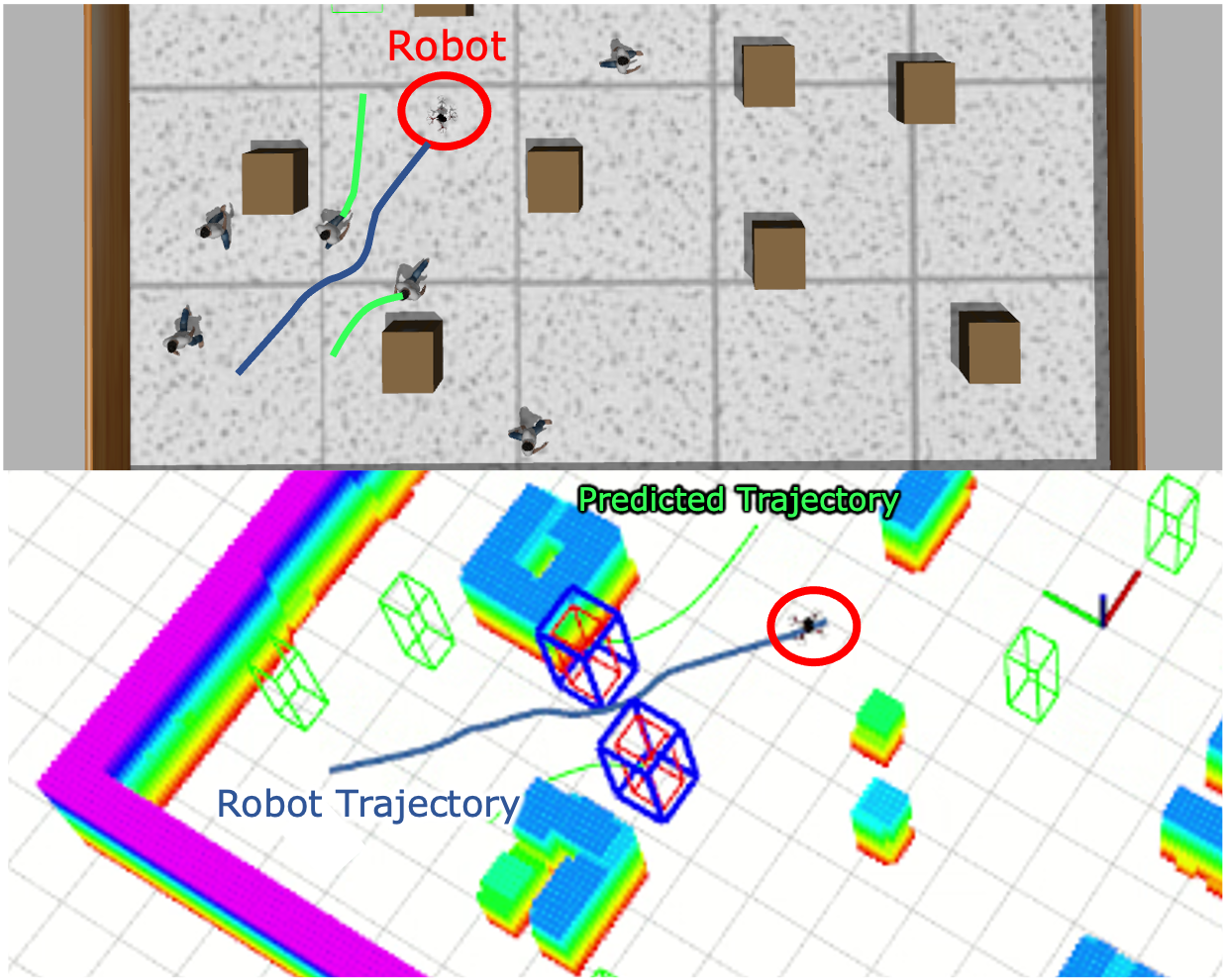}
    \caption{Visualization of obstacle tracking results in simulation. The green boxes are ground truth bounding boxes of obstacles. The red boxes are the ground truth obstacles in the camera FOV. The detected obstacles with predicted trajectories are shown as blue bounding boxes and green curves.}
    \label{fig:simulation}
\end{figure}
\textit{a) Simulation Experiments:} To evaluate the performance of the proposed system, we prepare three simulation environments with dynamic obstacles. An example of simulation environment experiments is shown in Fig. \ref{fig:simulation}. The environment consists of static obstacles with 14 pedestrians as the dynamic obstacles. In the bottom of Fig. \ref{fig:simulation}, we can see that the robot builds a static map for the environment, detects two dynamic obstacles marked as blue bounding boxes, and cleans their static map trails. The obstacles' trajectories are then predicted, and the quadcopter plans a path to navigate safely in the environment, shown as the blue curve.

\begin{figure}[t]
    \centering
    \includegraphics[scale=0.205]{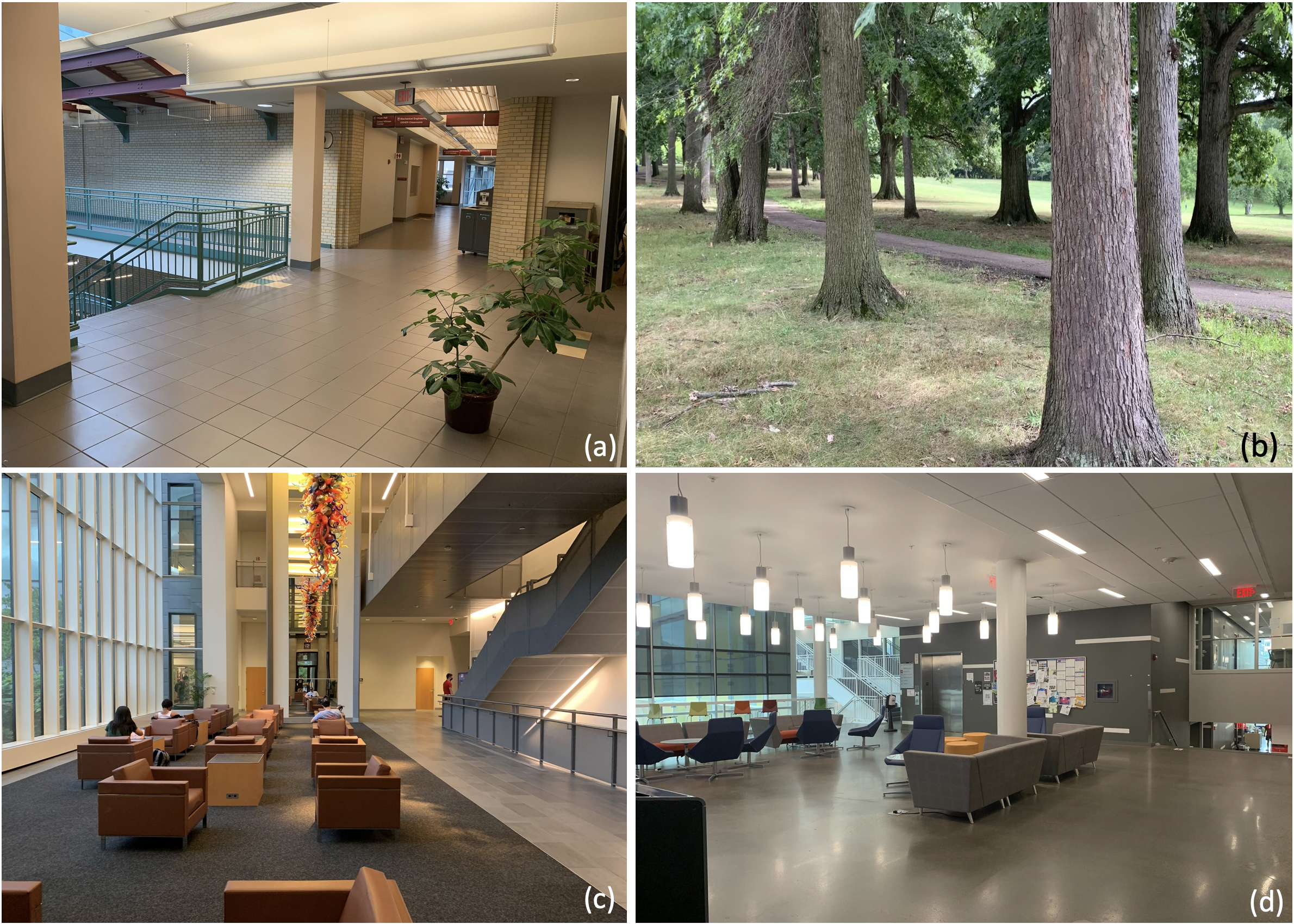}
    \caption{The physical test environments with pedestrians for evaluating our dynamic obstacle tracking and mapping system.}
    \label{fig:real_world_scenarios}
\end{figure}

\begin{figure*}[h]
    \centering
    \includegraphics[scale=0.55]{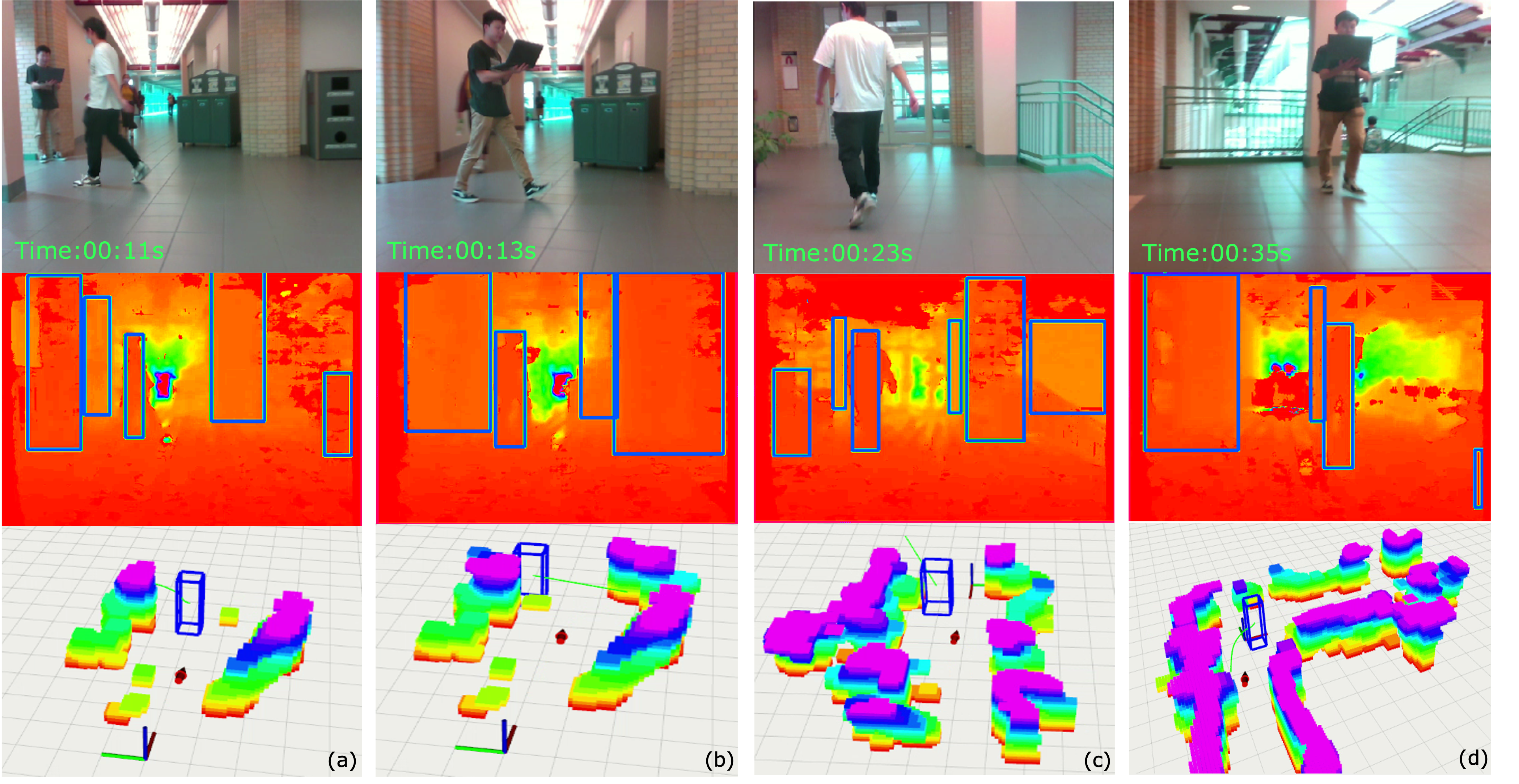}
    \caption{Visualization of the experiment in a L-shape corridor. The RGB images are shown in the upper figure, and the corresponding depth images with 2D obstacle detection results are shown in the middle. The final maps with dynamic obstacles are visualized in the bottom figure.}
    \label{fig:L_shape_result}
\end{figure*}

\textit{b) Physical Experiments:} The physical experiments are conducted in five different environments shown in Fig. \ref{intro_figure} and Fig. \ref{fig:real_world_scenarios}. We first test the proposed method, carrying the quadcopter by hand, in the four environments (Fig. \ref{fig:real_world_scenarios}). One example of dynamic obstacle tracking and mapping results is shown in Fig. \ref{fig:L_shape_result}. We can see that the walking pedestrians are detected and represented by the blue bounding boxes with the green lines indicating the predicted trajectories, while the person in static is not identified as the dynamic obstacle. Meanwhile, the static structures are captured by the static occupancy map with the dynamic obstacles' region cleaned. In addition, the trajectory prediction module can produce collision-free predicted trajectories considering the static obstacles. In Fig. \ref{fig:L_shape_result}c, the person is walking toward the corner of the L-shape corridor, so a left-turn trajectory is predicted, which matches the actual moving direction.

An indoor autonomous flight experiment is also performed to demonstrate the system's capability to avoid dynamic obstacles shown in Fig. \ref{intro_figure}. We can see that the quadcopter is able to detect a person walking toward it and map the static environment. With the trajectory planner mentioned in \cite{zhefan_new}, our system can successfully navigate dynamic environments.

\subsection{Benchmark Comparison}
To quantitatively analyze the system performance, we compare our method with the state-of-the-art dynamic obstacle detectors in the UAV platform \cite{Robust_avoidance_Umap}\cite{ZJU_dynamic_avoidance}. The comparison of dynamic obstacle detection errors in position and velocity is shown in Table \ref{table1}. In the physical experiment, the ground truth position is measured by the OptiTrack motion capture system. From Table \ref{table1}, we can see that our proposed method outperforms Method I in both velocity and position estimation accuracy. From our experiment observations, the worse performance of Method I is because it does not identify static and dynamic obstacles, resulting in incorrect estimations for static obstacles. The result also proves that our system has better velocity and comparable position estimation accuracy compared to Method II. Our proposed velocity estimation method treats dynamic obstacles as a whole, thus reducing noises from point cloud of other parts of the object and giving an accurate estimation.

\begin{table}[h]
    \begin{center}
    \caption{The benchmark of the detected dynamic obstacles' position and velocity errors in simulation and physical experiments.}
    \label{table1}
    \begin{tabular} { | c | c | c | c |}
    \hline
    Scenario & Method & Pos. Error (m) & Vel. Error (m/s)  \Tstrut\\
    \hline
    \multirow{3}{*}{\makecell [c] {Simulation \\Experiment}} & Method I \cite{Robust_avoidance_Umap} & 0.14 & 0.36 \Tstrut\\
    \cline{2-4}
    & Method II \cite{ZJU_dynamic_avoidance} & 0.11 & 0.19 \Tstrut\\ 
    \cline{2-4}
    & \textbf{Ours} & \textbf{0.11} & \textbf{0.08} \Tstrut\\

    \hline
   \multirow{3}{*}{\makecell [c] {Physical\\ Experiment}} & Method I \cite{Robust_avoidance_Umap} & 0.28 & 0.47 \Tstrut\\
   \cline{2-4}
   & Method II \cite{ZJU_dynamic_avoidance} & \textbf{0.18} & 0.29 \Tstrut\\
   \cline{2-4}
   & \textbf{Ours} & 0.19 & \textbf{0.21} \Tstrut\\

    \hline
    \end{tabular}
    \end{center}

\end{table}

The runtime evaluation of the proposed system on the Nvidia Xavier NX onboard computer is shown in Table \ref{table2}. Overall, the entire system takes less than 40ms in each iteration and can run over 25Hz. The region proposal detection contributes the majority of computation time. Besides, we also compare our method with the learning-based detection method, YOLO, on the onboard computer. The running time of YOLO is 256.4ms when running with the visual-inertial odometry, which is significantly slower than our system and is not able to handle real-time navigation.
\begin{table}[h]
    \centering
    \caption{The runtime of each module of the proposed system.}
    \begin{tabular}{ c c c } 
    \hline
    Modules & Time (ms) & Portion (\%) \Tstrut\\
    \hline
    Region Proposal Detection & 38.65 & 97.89\% \Tstrut\\ 
    Identification and Tracking & 0.26 & 0.66\% \\ 
    Trajectory Prediction & 0.57 & 1.45\% \\ 
    \textbf{System Total} & 39.49 & 100.00\% \\
    \hline

    \end{tabular}
    \label{table2}
\end{table}

\begin{figure}[t]
    \centering
    \includegraphics[scale=0.26]{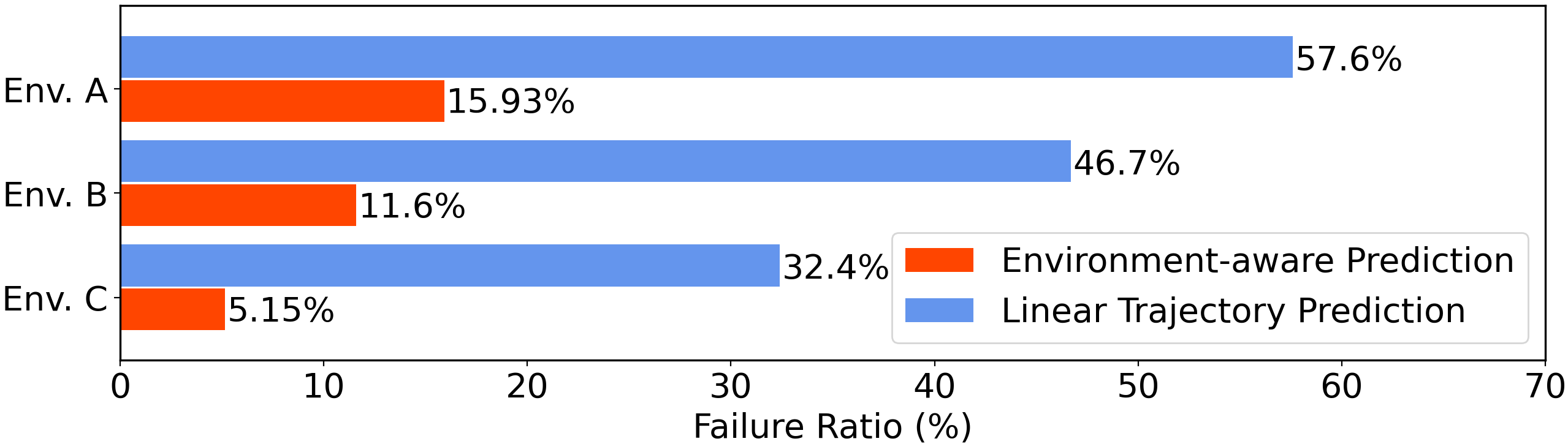}
    \caption{Comparison of the failure ratio between our proposed method and the linear predictor in three different simulation environments.}
    \label{fig:histogram}
\end{figure}

To evaluate the performance of the proposed environment-aware trajectory prediction module, we collect all predicted trajectories in three simulation environments and count the number of failed predictions. The number of the prediction paths in the library is set to 5. The failure prediction denotes either the trajectory is not collision-free or the predictor cannot find a solution. The comparison of the failure ratio between our proposed method and the linear trajectory prediction is shown in Fig. \ref{fig:histogram}. 
Among the three environments, the Env. A is the most complicated one, while the Env. C is the simplest one. It is shown that our prediction method still has a much lower failure ratio compared to the linear prediction in three environments. Besides, when the environment becomes more complex, the linear prediction's failure ratio goes up dramatically, while our method can always have a low value. For our methods, we can add the number of paths in the library to decrease the failure ratio. 

%
%

\section{Conclusion and Future Work}
This paper presents a novel dynamic obstacle tracking and mapping system for autonomous UAV navigation. The proposed method uses a 3D hybrid map, utilizing the occupancy voxel map to represent static obstacles and track dynamic obstacles as bounding boxes. The proposal detector module obtains the region proposals for obstacles from the depth image, which is refined by the proposed map refinement. Then, the proposed identification and filtering methods are applied to track dynamic obstacles. Besides, this work introduces a novel dynamic obstacle trajectory prediction algorithm based on the Markov chain, which considers the trajectory interaction with static environments. The results show that our dynamic obstacle detection method has low estimation errors in position and velocity and can be used to navigate dynamic environments safely. In the future, it is promising to explore camera models with a larger field of view to track dynamic obstacles.

\section{Acknowledgement}
\noindent The authors would like to thank TOPRISE CO., LTD and Obayashi Corporation for their financial support in this work.

\bibliographystyle{IEEEtran}
\bibliography{bibliography.bib}

\end{document}